\documentclass[letterpaper, 10 pt, conference]{ieeeconf} 

\IEEEoverridecommandlockouts                              
\usepackage{graphics} % for pdf, bitmapped graphics files
\usepackage{epsfig} % for postscript graphics files
\usepackage{mathptmx} % assumes new font selection scheme installed
\usepackage{times} % assumes new font selection scheme installed
\usepackage{amsmath} % assumes amsmath package installed
\usepackage{amssymb}  % assumes amsmath package installed
\usepackage[dvipsnames]{xcolor}
\usepackage[utf8]{inputenc}
\usepackage{pdfpages}
\usepackage{paralist}
\usepackage{array}
\usepackage{booktabs}
\usepackage{lipsum}
\usepackage{siunitx}
\usepackage{cite}
\usepackage{mathtools}
\usepackage{xcolor}
\usepackage{blkarray, bigstrut}
\usepackage{svg}
\let\labelindent\relax
\usepackage{enumitem}
\usepackage[bookmarksopen, bookmarksdepth=2, breaklinks=true ]{hyperref}
\RequirePackage{color,graphicx}
\definecolor{linkcolour}{rgb}{0.2,0.285,0.918}
\hypersetup{colorlinks,linkcolor={blue},citecolor={linkcolour},urlcolor={blue}} 
\usepackage{graphicx}

\title{{\LARGE \bf \textit{Adversarial Agent Behavior Learning in Autonomous Driving Using Deep Reinforcement Learning.\\ }} {\Large }}\vspace{-15pt}
\author{Arjun Srinivasan, Anubhav Paras and Aniket Bera\\
{University of Maryland, College Park, USA}\\
}
\newcolumntype{M}[1]{>{\centering\arraybackslash}m{#1}}

\newcommand\Item[1][]{%
  \ifx\relax#1\relax  \item \else \item[#1] \fi
  \abovedisplayskip=0pt\abovedisplayshortskip=0pt~\vspace*{-\baselineskip}}

\begin{document}

\maketitle

\begin{abstract}
Existing approaches in reinforcement learning train an agent to learn desired optimal behavior in an environment with rule based surrounding agents. In safety critical applications such as autonomous driving it is crucial that the rule based agents are modelled properly. Several behavior modelling strategies and IDM models are used currently to model the surrounding agents. We present a learning based method to derive the adversarial behavior for the rule based agents to cause failure scenarios. We evaluate our adversarial agent against all the rule based agents and show the decrease in cumulative reward.
\end{abstract}

%%%%%%%%%%%%%%%%%%%%%%%%%%%%%%%%%%%%%%%%%%%%%%%%%%%%%%%%%%%%%%%%%%%%%%%%%%%%%%%%
\section{Introduction}

Decision-making for autonomous vehicles is a higher-level planning problem that takes into account the behavior of other vehicles, pedestrians, autonomous vehicles in the environment of the ego vehicle. The decision-making algorithm in safety-critical applications such as autonomous driving need to work properly, otherwise it has resulted in many accidents in the past. Google’s self-driving car had a collision with an oncoming bus during a lane change \cite{google} which was due to incorrect assumption of the surrounding agents behavior. Rule-based approaches mostly result in conflicts as the assumptions made by any model deployed do not factor all the possible situations. 

Reinforcement learning (RL) has seen success in areas such as learning to play Atari Games \cite{atari}, and Go \cite{alphago} by DeepMind which has attracted many researchers to apply it in robotics \cite{robrl} and autonomous driving field \cite{rlav1,rlav2}. The drawback with the existing methods is that most of the environments for the RL training is rule based. As it is the case in autonomous driving algorithms too, there need some kind of adversarial agents in place of rule based agents to simulate situations which rule based agents do not cover. This requires us to train an adversarial agent in presence of a existing ego-agent which is trained in a rule based environment. There are many challenges associated with training such agents which we try to address in our work.

We propose adversarial reward formulation which addresses the problem of creating failure scenarios for the rule based agents. The autonomous driving scenario gives rise to the problem of multi agent interactions and self play associated with it. We develop a framework for single and multi agent training for generating adversarial policies for the surrounding agents. The success of our proposed scheme is verified by evaluating the existing trained agents in our adversarial environment. We also train a robust policy for ego agent to overcome the adversaries generated by our framework. We make use of the current state-of-the art RL algorithm Proximal Policy Optimization to generate the adversarial Policy.

\begin{figure}[t]
    \centering
    \includegraphics[width=1\linewidth]{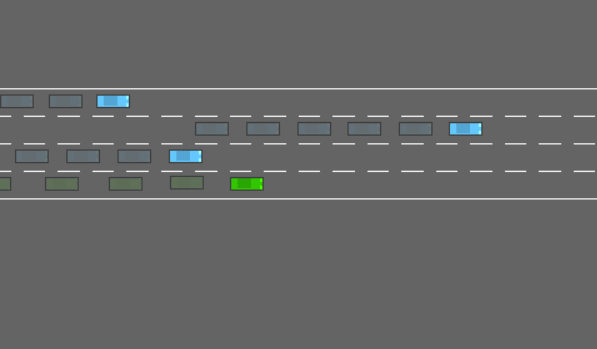}
    \caption{\textit{The green box denotes the ego-agent. The  blue boxes are the surrounding rule based agents in the highway environment. The traces of the boxes denote the trajectory of the agents.  } }
    \label{fig:Inroduction}
\end{figure}

The key contributions of our work can be summarized as follows:
\begin{enumerate}[nolistsep]
    \item We present a method to train adversarial behavior of autonomous driving agents through an adversarial reward formulation.
    \item We evaluate our adversarial agent with the cooperative multi-agent policy and verify the degradation in performance.
    \item We provide a defense pipeline against the attacks of adversarial agents and also compare the performance in standard environments.
\end{enumerate}
%%%%%%%%%%%%%%%%%%%%%%%%%%%%%%%%%%%%%%%%%%%%%%%%%%%%%%%%%%%%%%%%%%%%%%%%%%%%%%%%
\section{Related Work}
\label{sec:Related}

\subsection{Reinforcement Learning in Autonomous Driving}
 In the RL paradigm an autonomous agent learns to improve its performance at an
assigned task by interacting with its environment without explicit commands. RL is being used for developing high level driving policies for scenario based driving tasks in highways such as intersections, merges, roundabouts, lane keeping and overtaking.
Sallab et al. \cite{rlad1} propose a Deep RL system using DQNs and Deep Deterministic Actor Critic (DDAC) \cite{ddpg} to follow lane and maximize velocity. Wang et al. \cite {rlad3} propose an architecture using LSTM to model temporal dependencies to ramp merge the ego vehicle into a highway.

Sallab et al.\cite{drlsurvey} has a brief summary of a Deep reinforcement learning methods that are currently being used and derives a novel framework that uses RNNs to integrate information of the states and Deep Q networks/Deep Deterministic Actor Critic methods (for discrete/continuous action space) to plan the next action based on the highest reward. Chen et al.\cite{mfree} propose a rare approach that addresses the task of maneuver in roundabouts in a lighter computational load. This is achieved with a bird view representation with the use of three RL algorithms like DDQN, TD3, and Soft actor critic method that are employed with some tricks. The major drawback in all these approaches are that they train the ego agent with a rule based surrounding agents. We propose a method using RL approach to get an adversarial behavior of the surrounding agents.

\subsection{Multi-Agent Reinforcement Learning}
The presence of multiple agents in the environment demands an extension to Multi agent Reinforcement Learning (MARL) as there can be various scenarios such as cooperative, competitive and mixed behaviors among agents. Palanisamy et. al \cite{cad} propose a multi agent learning system for autonomous vehicles and provides a simulation setup for training. Lowe et. al \cite{mixcoop} devise an actor-critic approach for multi-agent learning in mixed cooperative environments.The multi-agent CAD problem in \cite{palanisamy} has been uniquely solved using Partially Observable Stochastic Games (POSG). Their contributions also involve the development of a novel multi-agent Gym environment that enables research of policies in different complicated scenarios. The presented results include the training and evaluation of one such scene of non-communicating similar vehicles with partial observability in an urban intersection that closely depicts a real world situation. 

We make use of the Open-AI based gym environment \cite{highway-env}  and modify it to suit the problem setup. The problem that arises during multi-agent adversarial training is the reward allocation. We need to identify the action among different adversarial agents in the setup and identify the action of the agents that causes higher reward. Motivated by works of Nguyen et al. \cite{marlci2}, we device our contribution identification method and train adversarial agent using TD3 method which is an improvement over DDPG method. We save the replay buffers and use the episodes that cause failure of the ego agent and create balanced data distributions for training.
%framework that should work well in this environment post successful evaluation from a 2D simulator like BARK/? for a multi agent following/overtaking problem.

\subsection{Adversarial  Learning}
Adversarial learning in reinforcement learning involves getting a robust policy to model initialization, disturbances in the environment. Modelling adversaries in computer vision tasks are much simpler compared to decision making tasks escpecially which are safety critical. Szegedy et al. \cite{sz} worked on designing  adversarial images that make an image classifier work improperly. Pinto et al. \cite{pinto} propose a method to train an adversarial agent for modeling disturbances in zero-sum game.
These kinds of methods can't be directly applied to autonomous driving problem, as the vehicles are not fully competitive and can break the zero-sum assumption. We propose an adversarial reward formulation which has a personal reward to maintain its own path and adversarial reward for making the ego-agent crash.

%%%%%%%%%%%%%%%%%%%%%%%%%%%%%%%%%%%%%%%%%%%%%%%%%%%%%%%%%%%%%%%%%%%%%%%%%%%%%%%%

\begin{figure}[t]
    \centering
    \includegraphics[width=1\linewidth]{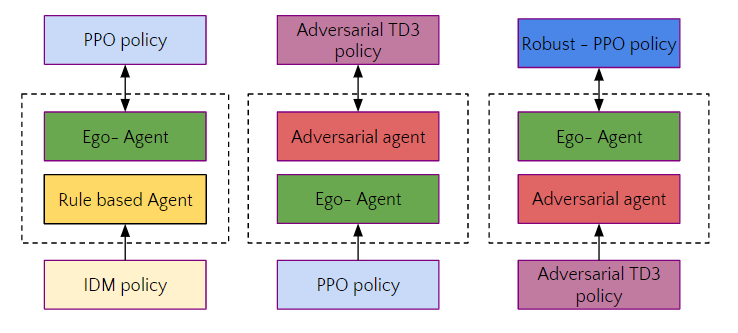}
    \caption{\textit{Overview of training process} }
    \label{fig:intro}
\end{figure}
\section{Overview}
\label{sec: Overview}

\subsection{Proximal Policy Optimisation - PPO}
To update the policy for the ego-agent, we chose the Proximal Policy Optimization (PPO) algorithm \cite{ppo}, which is a type of policy gradient reinforcement learning algorithm.

Policy gradient methods target modeling and optimizing the policy directly. The policy is usually modeled with a parameterized function with respect to $\theta$, denoted as $\pi_{\theta}(a|s)$.
Policy gradient methods maximize the expected total reward by repeatedly estimating the gradient, which has the form:
\[
\nabla J_{\theta}(\pi_{\theta}) = \mathbb{E}\left[ \sum_{t = 0}^{\infty} \psi_t \nabla_{\theta} \log \pi_{\theta} (a_t | s_t) \right]
\]

Here, $\psi_t$ can be the state-action value function ($Q^{\pi}(s_t,a_t)$), the advantage function ($A^{\pi}(s_t,a_t)$), the total reward function, or the reward following action $a_t$. Taking a step with this gradient pushes up the log-probabilities of each action in proportion to $\psi_t$.

PPO bounds the update of the policy to a trust region by using a less complex clipping method compared to Trust Region Policy Optimization (TRPO) \cite{trpo}. This ensures that the policy does not diverge between two consecutive training iterations. At every time step, it tries to minimize the cost function without deviating too much from the previous one. The objective function used in PPO is given by:
\[
L^{CLIP}(\theta)=\hat{\mathbb{E}}_{t}\left[\min \left(r_{t}(\theta) \hat{A}_{t}, \operatorname{clip}\left(r_{t}(\theta), 1-\varepsilon, 1+\varepsilon\right) \hat{A}_{t}\right)\right]
\]
where $\theta$ is the policy parameter, $\hat{\mathbb{E}}_{t}$ is the expectation over $t$ time steps, $r_{t}$ is the ratio of probability under the new and old policies, $\hat{A}_{t}$ is the estimated advantage at time $t$, and $\varepsilon$ is a hyperparameter typically set to 0.1 or 0.2. This clipping strategy is efficient and avoids complex KL divergence calculations. PPO is an on-policy method that supports continuous as well as discrete action spaces.

% \subsection{Twin Delayed DDPG - TD3}

% TD3 is an off-policy algorithm and similar to DDPG uses an experience replay buffer. Replay buffer is a set of previous experiences. We used TD3 to train our adversarial agent. Replay buffer stabilizes the learning of Q-function. \\

% The Q-learning algorithm is commonly known to suffer from the overestimation of the value function. This overestimation can propagate through the training iterations and negatively affect the policy. This property directly motivated Double Q-learning \cite{double_ql} and Double DQN \cite{double_dqn}: the action selection and Q-value update are decoupled by using two value networks. \\
% Twin Delayed Deep Deterministic (short for TD3; Fujimoto et al., 2018 \cite{td3_Fujimoto}) applied a couple of tricks on DDPG\cite{ddpg} to prevent the overestimation of the value function: \\
% \begin{enumerate}[wide, labelwidth=!, labelindent=0pt]
%     \itemsep 0.5em
%     \item{\textit{Clipped Double Q-Learning}}:
%     TD3 learns two Q-functions instead of one (hence “twin”), and uses the smaller of the two Q-values to form the targets in the Bellman error loss functions.
    
%     \item{\textit{Delayed update of Target and Policy Networks}}:
%     TD3 updates the policy (and target networks) less frequently than the Q-function. The paper recommends one policy update for every two Q-function updates.
    
%     \item{\textit{Target Policy Smoothing}}: TD3 adds noise to the target action, to make it harder for the policy to exploit Q-function errors by smoothing out Q along changes in action.

% \end{enumerate}

\subsection{Twin Delayed DDPG - TD3}

TD3 is an off-policy algorithm similar to DDPG, which uses an experience replay buffer. The replay buffer stores a set of previous experiences. We employed TD3 to train our adversarial agent, as the replay buffer stabilizes the learning of the Q-function.

The Q-learning algorithm is commonly known to suffer from the overestimation of the value function. This overestimation can propagate through the training iterations and negatively affect the policy. This property directly motivated Double Q-learning \cite{double_ql} and Double DQN \cite{double_dqn}: the action selection and Q-value update are decoupled by using two value networks.

Twin Delayed Deep Deterministic Policy Gradient (TD3; Fujimoto et al., 2018 \cite{td3_Fujimoto}) applied a couple of tricks on DDPG \cite{ddpg} to prevent the overestimation of the value function:

% \begin{enumerate}
%     % \itemsep 0.5em
%     \item{\textit{Clipped Double Q-Learning}}:
%     TD3 learns two Q-functions instead of one (hence “twin”) and uses the smaller of the two Q-values to form the targets in the Bellman error loss functions.
    
%     \item{\textit{Delayed Update of Target and Policy Networks}}:
%     TD3 updates the policy (and target networks) less frequently than the Q-function. The paper recommends one policy update for every two Q-function updates.
    
%     \item{\textit{Target Policy Smoothing}}: TD3 adds noise to the target action to make it harder for the policy to exploit Q-function errors by smoothing out Q along changes in action.
% \end{enumerate}
\begin{enumerate}
    \item \textit{Clipped Double Q-Learning}:
    % TD3 learns two Q-functions instead of one (hence “twin”) and uses the smaller of
    % the two Q-values to form the targets in the Bellman error loss functions.
    \item \textit{Delayed Update of Target and Policy Networks}:
    TD3 updates the policy (and target networks) less frequently than the Q-function. The paper recommends one policy update for every two Q-function updates.
    \item \textit{Target Policy Smoothing}: TD3 adds noise to the target action to make it harder for the policy to exploit Q-function errors by smoothing out Q along changes in action.
\end{enumerate}

\subsection{Behavior modelling for the rule based surrounding agents}
 \label{idm} 
The lateral and longitudinal behaviors of the vehicles are similar to rule based IDM vehicle and are derived from  \cite{latidm,longidm}. The longitudinal dynamics are derived from these equations.
$$
\begin{array}{l}
\dot{v}=a\left[1-\left(\frac{v}{v_{0}}\right)^{\delta}-\left(\frac{d^{*}}{d}\right)^{2}\right] \\
\\
d^{*}=d_{0}+T v+\frac{v \Delta v}{2 \sqrt{a b}}
\end{array}
$$
Where $v$ is the vehicle velocity, $d$ is the distance to its front vehicle, $v_{0}$ is the desired velocity, as target velocity, $T$ is the desired time gap, $d_{0}$ the jam distance, $a,b$ the maximum acceleration and deceleration, $\delta$ the velocity exponent. Longitudinal model computes an acceleration given the preceding vehicle’s distance and speed. Lateral model decides when to change lane by maximizing the acceleration of nearby vehicles. The following equations decide lane change lateral behavior. Condition for safe cut-in is:
$$
\tilde{a}_{n} \geq-b_{\text {safe }} \text { ; }
$$
If there is an incentive for the ego-vehicle and its followers the cut-in equation is:
$$
\underbrace{\tilde{a}_{c}-a_{c}}_{\text {ego-vehicle }}+p(\underbrace{\tilde{a}_{n}-a_{n}}_{\text {new follower }}+\underbrace{\tilde{a}_{o}-a_{o}}_{\text {old follower }}) \geq \Delta a_{\mathrm{th}},
$$

Where $c$ is the center (ego-) vehicle, $o$ is its old follower before the lane change, and $n$ is its new follower after the lane change. $a, \tilde{ a}$ are the acceleration of the vehicles before and after the lane change, respectively. $p$ is a politeness coefficient. $\Delta a_{\mathrm{th}}$ is the acceleration gain required to trigger a lane change. $b_{safe}$ is the maximum braking imposed to a vehicle during a cut-in.

%%%%%%%%%%%%%%%%%%%%%%%%%%%%%%%%%%%%%%%%%%%%%%%%%%%%%%%%%%%%%%%%%%%%%%%%%%%%%%%%
\section{Method}
\label{sec: Method}

\subsection{Training the adversarial agent }
The training for our proposed method is divided into three steps as follows:
\begin{enumerate}[nolistsep]
    \item An environment with ego-agent surrounded by rule based agents according to IDM policy as described in \ref{idm} . We train the ego-agent and get the PPO-policy.
    \item We train the adversarial agent according to adversarial reward formulation using TD3 algorithm. The surrounding agent is according to PPO-policy derived in Step 1. We derive the Adversarial TD3 policy.
    \item The environment consists of adversarial agents trained in Step 2. We train an ego-agent using PPO to derive a Robust-PPO policy to overcome the adversarial situations created by the agents.
\end{enumerate}
The above steps are represented in a block wise manner in Figure \ref{fig:intro}. After Step-2 of training we validate our trained Adversarial TD3 policy by observing a decrease in rewards for the ego agents PPO-policy from Step-1. Then we evaluate the improvement in performance with the Robust-PPO policy from Step 3 in an environment with our adversarial agents.

\subsection{Observation and Action Spaces}
The simulation environment that we set up consists of a bird's eye view of a highway with 4 as the chosen number of lanes. The surrounding agents have a velocity based on their behavior policy. We have an option that for all environments, several types of observations can be used and can be defined in the observation module. Each environment comes with a default observation, which can be changed or customized using environment configurations. We specifically set up a Kinematic Observation space, which is an $V \times F$ array where V is the list of nearby vehicles and F is the set of their features.

% \begin{figure}[h]
%     \centering
%     \includegraphics[scale = 0.5, width = \linewidth]{tab.png}
%     \caption{The table here shows a sample observation space for 1 ego vehicle with 4 surrounding vehicles.}
%     \label{fig:directive}
% \end{figure}\\

The observation space for each agent is a 5 × 5 array that describes a list of 5 nearby vehicles by a set of features  such as position ($x, y$)  and velocity ($v_x, v_y$). For the purpose of multi-agent learning we concatenate the observations of the number of agents present into a tuple. 

Apart from the number of vehicles, we set the number of lanes to 4. Each episode duration is about 40 seconds. By default each of the features in $x, y, v_x, v_y$ is bounded. 
Now, the action is defined as a \textit{Discrete Meta Action}.  Such a type adds a layer of speed and steering control on top of the continuous low-level control. This ensures that the controlled ego-vehicle can automatically follow the road at any desired velocity. Hence, this gives us the choice of set points on the low-level control that we can place for the vehicle. The available meta-actions now consist of just changing the target lane and speed.

The full action space is defined as follows:
\begin{itemize}
    \item 0: LANE\_LEFT
    \item 1: IDLE
    \item 2: LANE\_RIGHT
    \item 3: FASTER
    \item 4: SLOWER
\end{itemize}
Here, the priority of each of these actions is the numerical value corresponding to it. At some scenarios, there is a possibility that a particular action is not feasible i.e for cases like when accelerating/decelerating beyond the maximum/minimum velocity. Hence, the environment has an additional feature which enables us to take the best possible action. Taking an unavailable action is equivalent to taking the IDLE action.

Similar to continuous actions, speed changes and lane changes are a parameter that can be disabled. Speed change, which is longitudinal, and lane change i.e lateral actions are used in our setting. 

\subsection{Reward Formulation}
We formulate different types of rewards to motivate the training agent to achieve its desired state. For the normal ego-agent we define \textit{collision penalty, high speed reward} and \textit{right lane reward}. For adversarial agent we have a different reward formulation such that the adversarial agent needs to follow lanes in highway and cause ego-agent to fail without it getting collided.

% \begin{enumerate}[wide, labelwidth=!, labelindent=0pt]
\begin{enumerate}
    \itemsep 0.5em
    \item \textbf{Collision Penalty} ($\mathbf{r_{c}}$):
    This is a negative reward to penalize the vehicle when it could decelerate or change lane to avoid a collision, but chooses otherwise to be in a collision with another agent or with a non-intelligent vehicle. It is defined as:
        \begin{align}
        \mathbf{r_{c}} = 
        \begin{dcases}
    -1, & \:\:if\:\:r_{collision}\:=\:\:true \\
    0, &\:\:if\:\:r_{collision}\:=\:\:false 
        \end{dcases} \notag
            % if \:\: D_{h} &= True \:\: and \:\: N_{i} \notin [I_{min}, I_{max}] \:\: and \:\: \mathcal{A} \neq I:\\
        % r_{i} &= -100 \\
        % \text{else if } D_{h} &= False \text{ and } \mathcal{A} = I: \\
        % r_{i} &= -100 
        \end{align}
    
    $r_{c}$ occurs when the vehicle has the opportunity to avoid the incoming vehicle (i.e., change lane, decelerate), has so far avoid too little (i.e., stay in lane, maintain speed), or is still choosing not to avoid (i.e., change to obstacle's lane, accelerate towards obstacle).
    
    \item \textbf{High Speed Reward} ($\mathbf{r_{s}}$):
    We give a positive reward when the agent vehicle accelerates and/or safely maintains a high speed, and a zero reward otherwise. It is formulated as: 
        \begin{align}
        \mathbf{r_{s}} = 
        \begin{dcases}
        0.5, & \:\:if\:\:accelerating\:\:or\:\:at\:\:max\:\:speed \\
        0, &\:\:if\:\:decelerating\:\:or\:\:low\:\:speed 
        \end{dcases} \notag
        \end{align}
    Before crafting this reward function, we have tried to use a reward function that gives reward for every second that the agent vehicle maintain a positive speed. However, although that reward function can teach the reinforcement learning policy to avoid collision, it does not encourage any high speed driving and the agent vehicle is becoming too conservative. We have found out giving reward for the agent to accelerate or maintain a high speed eliminates the side effects while also encourage the agent vehicles to pass any many other vehicles as possible.

    \item \textbf{Right Lane Reward} ($\mathbf{r_{r}}$):
    This reward function gives reward when the agent vehicle is driving on the right-most lane. The agent vehicle keeps getting this reward when staying on the right-most lane as it is defined as the high speed lane here. Agent vehicle staying on all other lanes except for the right-most lane will map to a reward of zero.
        \begin{align}
        \mathbf{r_{r}} = 
        \begin{dcases}
        0.3, & \:\:agent\:\:stays\:\:in\:\:right\:\:lane \\
        0, &\:\:otherwise 
        \end{dcases} \notag
        \end{align}
        $r_{r}$ is a constant reward for every timestamp that the agent vehicle stays in the right-most lane. Although $r_{r}$ reward value is not large, it is necessary for two reasons. First, if no such reward is implemented, the agent vehicle will keep changing between two safe lanes because of the previous reward function $r_{l}$. This behaviour is undesirable and does not make sense in real life. Second, the fastest driving vehicle staying on one side of the road (e.g. right-most or left-most lanes depends on regions) is the convention people follow in real life.

            \item \textbf{Adversarial Reward} ($\mathbf{r_{l}}$):
    We would like the adversarial agent to cause the ego-agent to fail but at the same time it should also navigate properly. So if the adversarial agent causes the ego-agent without This is formulated as: 
        \begin{align}
        \mathbf{r_{adv}} = 
        \begin{dcases}
        1, & \:\:only\:\: ego-agent\:\:collides\:\\
        0.5, & \:\:\: ego\:\: and\:\:  adversarial\:\:agent\:\:collides\:\\ -1, & \:\:only\:\: adversarial\:\:agent\:\:collides\:\\   
        0, &\:\:otherwise 
        \end{dcases} \notag
        \end{align}
    This kind of reward setting resulted in many cases where the trained ego-agent from PPO-policy to collide. 
\end{enumerate}

The total reward for ego-agent $\mathbf{r_{total}}$ here is a summation of all the rewards combined given by $\mathbf{r_{total}} = \mathbf{r_{c}} + \mathbf{r_{s}} + \mathbf{r_{r}}$. And for adversarial agent it is just $\mathbf{r_{adv}}$. In each episode, the reinforcement learning policy only look at this total reward and refine the policy to maximize it.

\begin{figure}[t]
    \centering
    \includegraphics[width=1\linewidth]{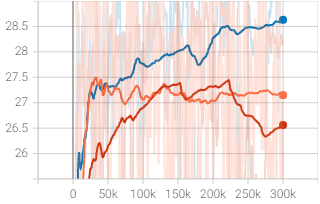}
    \caption{\textit{Training curves corresponding to single agent (red) training, adversarial agent (orange) training and robust agent (blue) training}. This indicates the higher reward obtaining capacity of our agents with the Robust PPO policy. }
    \label{fig:train}
\end{figure}

\begin{figure}[ht]
    \centering
    \includegraphics[scale = 0.5, width = \linewidth]{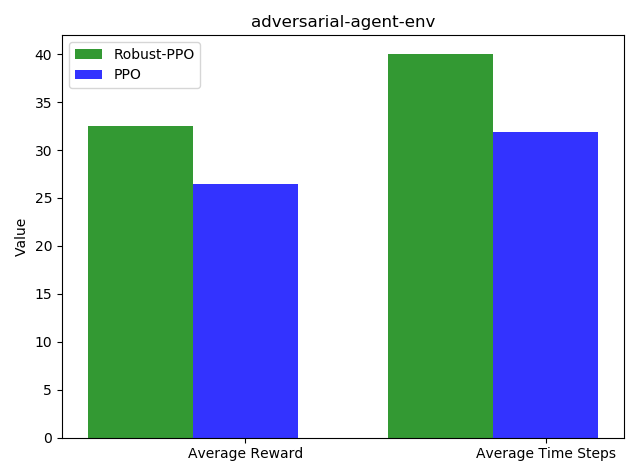}
    \caption{Evaluation of Robust-PPO policy in adversarial environment.}
    \label{fig:advenv}
\end{figure}
% \begin{figure}[h]
%     \centering
%     \includegraphics[scale = 0.5, width = \linewidth]{idm.png}
%     \caption{Evaluation of Robust-PPO policy in rule based IDM environment.}
%     \label{fig:directive}
% \end{figure}\\

\subsection{Evaluation of Adversarial Agent}
Our rewards values are modeled based on observations made in three stages of our experimentation. The training curves for the three steps of training are as shown in Fig \ref{fig:train}. First, we gave a high reward value to $r_{s}$, to make the vehicle driving as fast and far as possible to gain a greater reward. This caused the policy to only speed up the vehicle all the time, preventing the session to last due to fast collisions. In order to rectify this, we lower our $r_{s}$ to a much lower value.

In the final stage, the agent vehicles learn when to change lane to survive longer and increase its reward. However, it would stuck with infinite lane changing actions between two or three lanes after determining there are no vehicles in the side lanes. This hinders the policy to converge, so we introduced $r_{r}$ as a constant reward for driving in the right-most lane. Tuning $r_{r}$ to an optimal value led the agent vehicle to perform in a reasonable behavior pattern.
 
 The adversarial reward $r_{adv}$ for making the ego-agent train worked well, and this was evaluated with the decrease in reward obtained by the ego-agent following the PPO-policy from the first training step. Our Robust-PPO policy was able to get better average rewards and was able to prolong for more steps in the simulation. We ran the simulation for 100 times and averaged the time steps it ran, and the reward obtained, and tested in a 3 lane highway environment with adversarial surrounding agents, see Fig \ref{fig:advenv}.

\section{Conclusion}
We conduct experiments to analyse the behavior of ego agent under rule based and our simulated adversarial agents. We used an Open-AI based simulator “Highway-Env” and extend it to account for adversarial-agent learning. Our agents were able to  maximise the expected reward and converge well in the cases of rule based and adversarial environments. Our robust PPO policy trained in adversarial environment performed better compared to the one trained in rule based environment.

Our future goal is to utilise the reward allocation scheme to train the adversarial agent in a multi-agent scenario. We also aim to train for adversarial behaviors under various scenarios such as roundabouts, lane merging and 4-way intersections. We plan to introduce more robustness in the simulation environment as the current gym-environment doesn't provide different type of vehicle behaviors such as large trucks, motorcycles, bikes, etc.

%%%%%%%%%%%%%%%%%%%%%%%%%%%%%%%%%%%%%%%%%%%%%%%%%%%%%%%%%%%%%%%%%%%%%%%%%%%%%%%%

%%%%%%%%%%%%%%%%%%%%%%%%%%%%%%%%%%%%%%%%%%%%%%%%%%%%%%%%%%%%%%%%%%%%%%%%%%%%%%%%

%%%%%%%%%%%%%%%%%%%%%%%%%%%%%%%%%%%%%%%%%%%%%%%%%%%%%%%%%%%%%%%%%%%%%%%%%%%%%%%%
%  \addtolength{\textheight}{-8cm}   % This command serves to balance the column lengths
                                  
{\small
\bibliographystyle{IEEEtran}
\bibliography{root}

}

\end{document}